\documentclass[letterpaper, 10 pt, conference]{ieeeconf}  

\IEEEoverridecommandlockouts                              

\usepackage{cite}
\usepackage{xcolor}

\usepackage{amsthm}
\usepackage{amsmath,amssymb,amsfonts,amsthm}
\usepackage{algorithmic}
\usepackage{graphicx}
\usepackage{textcomp}
\usepackage{xcolor}
\usepackage{multicol}
\usepackage{multirow}
\usepackage{tabularray}
\usepackage{float}
\usepackage{subfigure}
\theoremstyle{definition}
\newtheorem*{theorem}{Theorem} 
\newtheorem*{remark}{Proof} 
\title{\LARGE \bf
An Analytic End-to-End Deep Learning Algorithm based on Collaborative Learning 
}

\author{Sitan LI and Chien Chern CHEAH 
}

\begin{document}

\maketitle
\thispagestyle{empty}
\pagestyle{empty}

\begin{abstract}
In most control applications, theoretical analysis of the systems is crucial in ensuring stability or convergence, so as to ensure safe and reliable operations and also to gain a better understanding of the systems for further developments. However, most current deep learning methods are black-box approaches that are more focused on empirical studies. Recently, some results have been obtained for convergence analysis of end-to end deep learning based on non-smooth ReLU activation functions, which may result in chattering for control tasks. This paper presents a convergence analysis for end-to-end deep learning of fully connected neural networks (FNN) with smooth activation functions. The proposed method therefore avoids any potential chattering problem, and it also does not easily lead to gradient vanishing problems. The proposed End-to-End algorithm trains multiple two-layer fully connected networks concurrently and collaborative learning can be used to further combine their strengths to improve accuracy. A classification case study based on fully connected networks and MNIST dataset was done to demonstrate the performance of the proposed approach. Then an online kinematics control task of a UR5e robot arm was performed to illustrate the regression approximation and online updating ability of our algorithm.

\end{abstract}
\begin{keywords}
End-to-End, Deep learning, Sigmoid, Robot kinematics
\end{keywords}
\section{Introduction}
Deep learning networks are widely applied in various applications due to their exceptional performance, but their use in control tasks is limited by the difficulty of analyzing the convergence. The commonly used methods for training Deep Neural Networks (DNNs) are backpropagation and gradient descent\cite{lecun2015deep}\cite{shrestha2019review}, but they are considered as black-box approaches that offer no assurance of convergence. In robot control, for example, convergence of error is crucial for ensuring that the robot moves accurately and smoothly. Otherwise, the robot may move erratically or fail to reach its intended target, which can be dangerous in industrial applications. Disregarding the theoretical convergence  of deep learning could hinder its progress as a dependable and trustworthy technology.
Therefore, it is crucial to develop an analytic learning algorithm for DNNs that can analyze convergence of the algorithm  and ensure the safe and robust deployment of deep learning. 

In recent years, researchers from machine learning field have started to analyze convergence of deep neural networks in optimization aspect.
 In \cite{zou2020gradient}, it was demonstrated that an over-parameterized deep neural network with rectified linear unit (ReLU) activation functions achieves global minima for training loss in a binary classification task. In \cite{du2019gradient}, it was proved that, with Gram Matrix structure and over-parameterized neural networks (NNs), training loss can converge to zero through GD. In \cite{allen2019convergence}, it analyzed over-parameterized DNNs and showed that stochastic gradient descent(GD) method is capable of finding a global minimum for non-smooth ReLU and also other smooth activation functions. 
However,  results in \cite{zou2020gradient}\cite{du2019gradient}\cite{allen2019convergence}  are based on assumptions  of over-parameterized networks
, which are hard to implement in real control systems. 

Layerwise learning is a promising approach for analyzing deep neural networks, whereby the network is divided into multiple layers or blocks and trained sequentially. \textcolor{black}{In \cite{shin2022effects}, the convergence analysis was provided for deep linear networks based on a layer-wise learning using block coordinate gradient descent. However, compared to deep nonlinear networks deep linear networks are not preferred   in practical scenarios as linear ones have poor approximation ability.  }In \cite{nguyen2022analytic},  an    analytical layer-wise approach was proposed for fully connected networks and was applied  to classification and robot kinematic control tasks. In \cite{nguyen2022layer},  a layer-wise deep learning algorithm with convergence analysis was proposed for convolutional neural networks. 

However, the implementation on layerwise learning  requires repeating procedure of adding one layer at a time, which is therefore limited to repetitive tasks.
For more general tasks, End-to-End deep learning methods are therefore required. In \cite{patil2021lyapunov},  real-time weight adaptation laws were developed based on  the Lyapunov-based stability analysis for a deep feed-forward neural network. However, it mainly focuses on the  control of dynamic systems and classification tasks are not considered. 
In \cite{10053845}, an End-to-End learning algorithm was developed  for DNNs for both image classification tasks and real-time control of robotic systems.
The convergence analysis was based on non-smooth ReLU activation function and its variants.

The non-smooth activation function ReLU are widely applied in existing end-to-end deep learning methods, but it is not differentiable when the input is zero.  This limits the implementation on control systems as it results in chattering of input. A \textcolor{black}{smooth activation function like sigmoid or Tahn }is more feasible as they are differentiable  at any point, which therefore does not result in any chattering problem in actual implementation. However, one main issue that limits the use of the \textcolor{black}{some smooth activation functions like Sigmoid activation function} in end-to-end deep learning is that the gradients can easily vanish when the magnitude of  input becomes   too large. This means that the derivative of the function becomes extremely close to 0, which can result in exponentially decreasing gradients as they propagate through the layers of the deep FNN.

In this paper, we proposed an  end-to-end deep learning  method with convergence analysis based on \textcolor{black}{smooth nonlinear activation functions}. \textcolor{black}{ The difficulty in this convergence analysis is due to the nonlinearities of the activation functions in the hidden layers.} Unlike existing end-to-end deep learning methods, the proposed learning method does not result in vanishing gradients easily, even when sigmoid activation functions are used. The proposed   learning method is developed based on the collaborative learning of several  sub-systems. The sub-systems are updated by a two-layer update law concurrently. We show that the sub-systems can also be combined  to form a more accurate and robust predictive model by leveraging the diversity and complementary strengths of the sub models using collaborative learning. It is a
powerful machine learning technique that involves 
multiple systems working together to achieve better accuracy than an individual system can achieve alone. \textcolor{black}{Collaborative learning \cite{wang2015collaborative}\cite{song2018collaborative} is a powerful machine learning technique that involves multiple systems working together to achieve better accuracy than an individual system can achieve alone. Some results have been obtained in the literature (see \cite{wang2015collaborative} and \cite{song2018collaborative} and the references therein) but the previous works do not provide any convergence analysis.} However, the previous works did not provide any convergence analysis. This paper presents a theoretical convergence analysis for the proposed end-to-end collaborative  system. To demonstrate the efficacy of the proposed learning algorithm, a case study was first done on a classification task based on fully connected networks and MNIST dataset. Then a regression problem was also performed on the kinematics of a UR5e robot arm.

\section{ End-to-End Deep learning with Collaborative learning}
\label{sec: problm_state} 
 
\begin{figure}[htbp]
    \centering
    \includegraphics[width=0.45\textwidth]{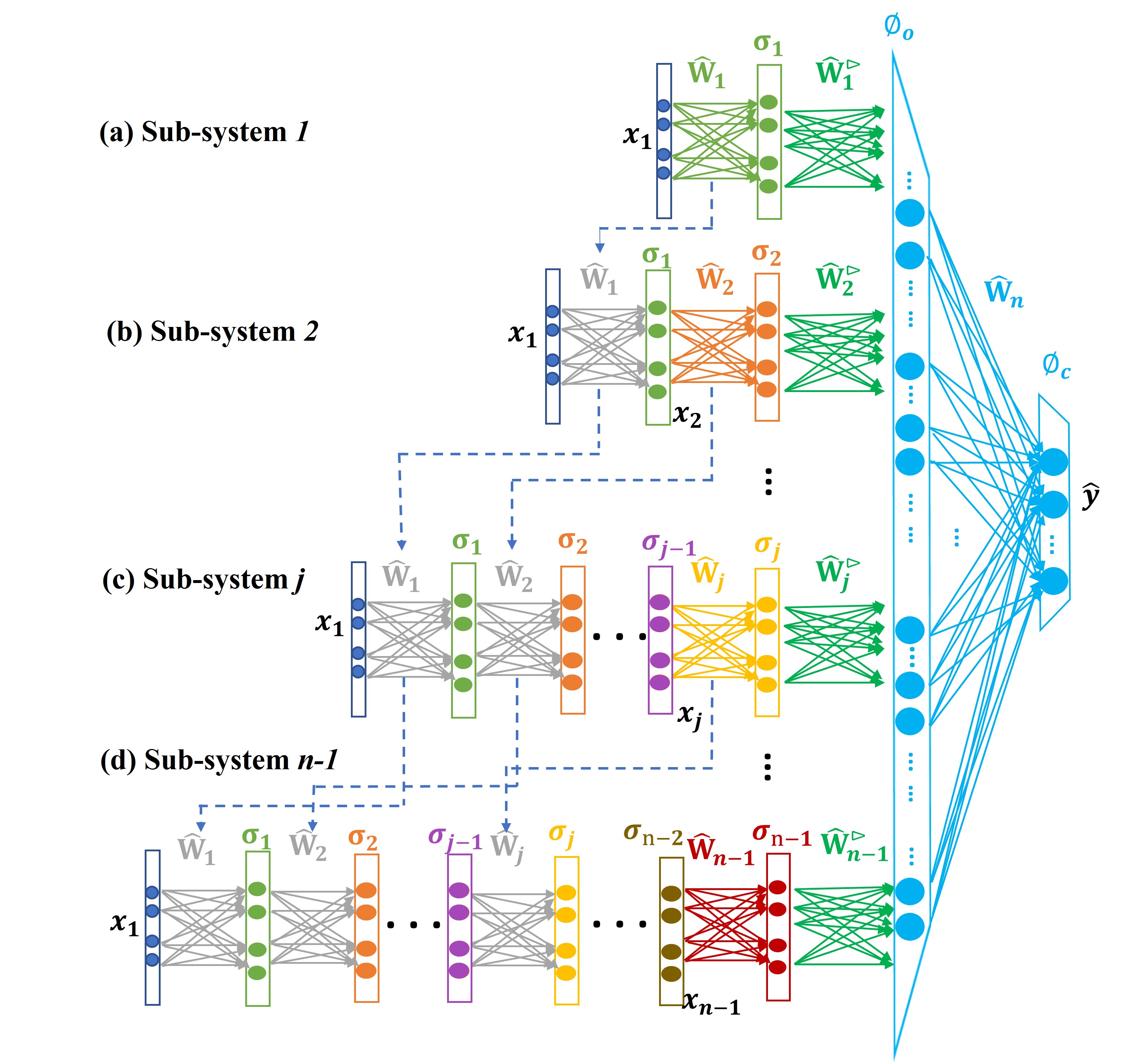}
    \caption{Collaborative End to End learning method, where the  collaborative FNN is combined of $n-1$  sub-systems}
    \label{fig:endtoend}
\end{figure}

A collaborative deep fully connected network is designed by combining \(n-1\) sub-systems  as shown in Fig \ref{fig:endtoend}. In the first  sub-system, the input goes through one hidden layer, as shown in Fig \ref{fig:endtoend}(a). The input weight matrix \({\mathbf{\hat W}}_{1}\) and the pseudo  weight matrix of  output layer \(\mathbf{\hat W}_1^{\vartriangleright}\) are updated together by the proposed updating law.  Fig \ref{fig:endtoend}(b) shows the learning in the second sub-system with weight matrix \({\mathbf{\hat W}}_{2}\) and pseudo  weight matrix \({\mathbf{\hat W}}_{2}^{\vartriangleright}\) . The input \(\boldsymbol{x}_2\) of the second sub-system is formed by passing \(\boldsymbol{x}_1\) through 
the first weight matrix \({\mathbf{\hat W}}_{1}\).  Fig \ref{fig:endtoend}(c) shows the updating in the $j$th sub-system with weight matrix \({\mathbf{\hat W}}_{j}\) and pseudo  weight matrix \({\mathbf{\hat W}}_{j}^{\vartriangleright}\).The learning of last weights takes place at $n-1$th sub-system as shown in Fig \ref{fig:endtoend}(d), where two  weights \({\mathbf{\hat W}}_{n-1}\) and \({\mathbf{\hat W}}_{n-1}^{\vartriangleright}\) are learned instead of pseudo weights. In this way, all weights  \(\mathbf{\hat W}_1, \mathbf{\hat W}_2, \dots, \mathbf{\hat W}_{n-1}\) are updated simultaneously  based on \(n-1\) sub-systems by the input data  \(\boldsymbol{x}(k)\). All $n-1$ sub-systems are connected to a fully connected layer to do the final classification. The weight matrix of the last fully connected layer $\mathbf{\hat W}_n$ is updated concurrently with the sub-systems.

After all the sub-systems have been updated using data $\boldsymbol x_1(k)$,  the weight matrix  \({\mathbf{\hat W}}_{1}\)  is shared with remaining subsystems as illustrated in  Fig \ref{fig:endtoend}, \({\mathbf{\hat W}}_{2}\)  is shared with remaining subsystems from 3 to $n-1$, and similarly, the weight matrix  \({\mathbf{\hat W}}_{j}\)  is shared with remaining subsystems from $j+1$ to $n-1$. The same updating procedure is then repeated for all sub-systems based on next data $\boldsymbol x_1(k+1)$.  When the entire set of data has been passed through the neural network,  one training epoch is finished. The sub-systems are trained for multiple epochs until it reaches convergence.  The  remaining part of this section presents the  update laws for simultaneously updating all weight matrices. \(\mathbf{\hat W}_1,...,\mathbf{\hat W}_n\).

For all   sub-systems, the input data is $\boldsymbol{x}_1({k})$. For ease of representation,  the input to the last 2 layers of $j$th sub system $\boldsymbol{x}_j({k})$, $j=2,..,n-1$, can be calculated by passing each input data $\boldsymbol {x}_1(k)$ through estimated weights \(\mathbf{\hat W}_1(k), \mathbf{\hat W}_2(k), \dots, \mathbf{\hat W}_{j-1}(k)\)  shared from previous sub-systems:
\begin{align}
\boldsymbol{x}_j({k}) = {\boldsymbol{\hat \sigma}}_{j-1}({k})  \label{eq:input_wj_FPL} 
\end{align}
\begin{equation}
\resizebox{.98\hsize}{!}{$
 {\boldsymbol{\sigma}}_{j-1}({k}) = \boldsymbol{\sigma}_{j-1}({\mathbf{\hat W}}_{j-1}({k})\boldsymbol{\sigma}_{j-2}( \dots  \boldsymbol{\sigma}_2({\mathbf{\hat W}}_2({k}) \boldsymbol{\sigma}_1({\mathbf{\hat W}}_1({k}) \boldsymbol{x}_1({k})))\dots)) $}
\end{equation}
 where     estimated weights are  denoted as \(\mathbf{\hat W}_1(k), \mathbf{\hat W}_2(k), \dots, \mathbf{\hat W}_{n}(k)\)   from first layer  to last layer of the $j$th sub-system and  \(\boldsymbol{\sigma}_j\) denotes the activation functions of the \(j\)th hidden layer, which can be \textcolor{black}{ any smooth and differentiable activation functions like Sigmoid activation function, Tahn activation function }.

For each subsystem, the last two weight matrices (see Fig \ref{fig:endtoend}) are updated by a two-layer update law. The estimated  weight matrices $\mathbf{\hat {W}}_j$ and  $\mathbf{\hat {W}}^{\vartriangleright}_j$  of   all   sub-systems and the weight matrix  $\mathbf{\hat {W}}_n$ of the last  layer are updated concurrently. 

In the $j$th sub-system,  the estimated output of this sub-system \(  {\boldsymbol{\hat y}_j}(k)\) is constructed by the estimated input weights  \(\hat{\mathbf{W}}_j(k)\) and output weights \(\hat{\mathbf{W}}_j^{\triangleright}(k)\) as follows:
\begin{equation}
\begin{aligned}
  {\boldsymbol{\hat y}_j}(k)
=&\boldsymbol{\phi}_{o}\big(\hat{\mathbf{W}}_j^{\triangleright}(k) \boldsymbol{\sigma}_{j}(\hat{\mathbf{W}}_j(k) \boldsymbol{x}_j(k))\big) \label{eq:est_yk_2}
\end{aligned}
\end{equation}
where the activation functions for output layer are denoted as \(\boldsymbol{\phi}_o\).
There exists ideal unknown weight matrices \(\mathbf{W}_j\) and \(\mathbf{W}_j^{\triangleright}\) for each epoch such that the actual output $\boldsymbol{y}(k)$ is represented as:
\begin{equation}
\begin{aligned}
\boldsymbol{y}(k)
=\boldsymbol{\phi}_{o}\big(\mathbf{W}_j^{\triangleright} \boldsymbol{\sigma}_{j}(\mathbf{W}_j \boldsymbol{x}_j(k))\big) 
\label{eq:yk_2}
\end{aligned}
\end{equation}

The estimated output of the collaborative multi-layer fully connected network (MLFN), which combines $n-1$ sub-systems is presented as follows: 
\begin{equation}
\begin{aligned}
\boldsymbol{\hat y}_{{n}}&(k) =\boldsymbol{\phi}_{c}(\mathbf{\hat W}_n(k) \boldsymbol{\hat \phi}_{sub}(k)) \label{eq:est_yk_n}
\end{aligned}
\end{equation}
 where $\boldsymbol{\hat \phi}_{sub}(k)= [\boldsymbol{\hat  y}_1(k),\boldsymbol{\hat  y}_2(k),..,\boldsymbol{\hat y}_j(k),..,\boldsymbol{\hat y}_n(k)]^T$, and the activation functions of the last layer of collaborative  network   are defined as \(\boldsymbol{\phi}_c\). There exists an ideal weight matrix \(\mathbf{W}_n\) for each epoch such that the ideal output of the collaborative network is represented as follows:
\begin{equation}
\begin{aligned}
\boldsymbol{y}(k)
=\boldsymbol{\phi}_{c}\big(\mathbf{W}_n\boldsymbol{\hat \phi}_{sub}(k)\big)
\label{eq:yk_n}
\end{aligned}
\end{equation}

The weight matrices \(\mathbf {\hat W}_j\) and \(\mathbf{\hat W}_j^{\triangleright}\) are then updated   by two learning laws based on data \(\boldsymbol{x}_j(k)\) and error. The   estimation error \(\boldsymbol{e}_j(k)\)  for the $j$th sub-system is defined as the difference between its estimated output in \eqref{eq:est_yk_2} and the ideal output \eqref{eq:yk_2} in as:  
\begin{align}
\boldsymbol{e}_j(k)
=&\boldsymbol{y}(k)-  \boldsymbol{\hat y}_j(k) \label{eq:err_2} \\
=&\resizebox{0.8\linewidth}{!}{$\boldsymbol{\phi}_{o}\big(\mathbf{W}_j^{\triangleright} \boldsymbol{\sigma}_{j}(\mathbf{W}_j \boldsymbol{x}_j(k))\big)    \boldsymbol{\phi}_{o}\big(\hat{\mathbf{W}}_j^{\triangleright}(k) \boldsymbol{\sigma}_{j}(\hat{\mathbf{W}}_j(k) \boldsymbol{x}_j(k))\big) $}\nonumber
\end{align}

For updating the last weight matrix $\mathbf{W}_n$, from \eqref{eq:est_yk_n} and \eqref{eq:yk_n} the error $\boldsymbol{e}_n(k)$ is formulated as:
\begin{equation}
\begin{aligned}
\boldsymbol{e}_n(k)
=\boldsymbol{\phi}_{c}\big( \mathbf{W}_n \boldsymbol{\hat \phi}_{sub}(k)\big)-\boldsymbol{\phi}_{c}\big(\mathbf{\hat W}_n (k)\boldsymbol{\hat \phi}_{sub}(k)\big)
\label{eq:err_nn}
\end{aligned}
\end{equation}

Let 
\begin{equation}
\boldsymbol{\delta}_j(k) = \mathbf{W}_j^{\triangleright} \boldsymbol{\sigma}_{j}(k) - \hat{\mathbf{W}}_j^{\triangleright}(k) \hat{\boldsymbol{ \sigma}}_{j}(k)\label{eq:delta_2}
\end{equation}  
where ${\boldsymbol{\sigma}}_{j}(k)=\boldsymbol{\sigma}_{j}({\mathbf{W}}_j \boldsymbol{x}_j(k))$ and $\hat{\boldsymbol{ \sigma}}_{j}(k) = \boldsymbol{\sigma}_{j}(\hat{\mathbf{ W}}_j(k) \boldsymbol{x}_j(k))$
, then equation \eqref{eq:delta_2}  can be expressed as:
\begin{equation} 
\begin{aligned}
\resizebox{.98\hsize}{!}{$\boldsymbol{\delta}_j(k) 
=\hat{\mathbf{W}}_j^{\triangleright}(k)\Delta \boldsymbol{\sigma}_{j}(k) + \Delta\mathbf{W}_j^{\triangleright}(k)\hat{\boldsymbol{\sigma}}_{j}(k){+\;} \Delta\mathbf{W}_j^{\triangleright}(k) \Delta \boldsymbol{\sigma}_{j}(k)$} \label{eq:deltak_2}
\end{aligned}
\end{equation}
where \(\Delta \boldsymbol{\sigma}_{j}(k) =  \boldsymbol{\sigma}_{j}(k) - \boldsymbol{\hat \sigma}_{j}(k)\) and \(\Delta\mathbf{W}_j^{\triangleright}(k) = {\mathbf{W}}_j^{\triangleright} -  \hat{\mathbf{W}}_j^{\triangleright}(k) \). In addition, let
\begin{equation}
\begin{aligned}
\resizebox{.97\hsize}{!}{$\boldsymbol{\delta}_n(k)
= \mathbf{W}_ n\boldsymbol{\hat \phi}_{sub}(k)-\mathbf{\hat W}_ n(k)\boldsymbol{\hat \phi}_{sub}(k)=\Delta\mathbf{W}_ n(k)\boldsymbol{\hat \phi}_{sub}(k) $}
\label{eq:deltak_n}
\end{aligned}
\end{equation}
where   \(\Delta\mathbf{W}_n(k) = {\mathbf{W}}_n -  \hat{\mathbf{W}}_n(k) \). 
The activation functions \(\boldsymbol{\phi}\) (including \(\boldsymbol{\phi}_o\) and \(\boldsymbol{\phi}_c\)) can be selected as  \textcolor{black}{monotonically increasing smooth activation functions with upper bounds like Sigmoid activation function, Tahn activation function} or Identity activation function. Since they are  monotonically increasing, they have the upper bounds $f_{\phi M }$.  The errors \(\boldsymbol{e}_j(k)\),\(\boldsymbol{e}_n(k)\) in \eqref{eq:err_2},\eqref{eq:err_nn} and \(\boldsymbol{\delta}_j(k)\),\(\boldsymbol{\delta}_n(k)\)  in \eqref{eq:deltak_2},\eqref{eq:deltak_n} have the following properties:
\begin{itemize}
	\item[i,]the sign of $i$th elements of \( \boldsymbol{e}_j(k)\)  and \(\boldsymbol{\delta}_j(k)\)  are the same, $j=1,2...,n$, i.e.
	\begin{equation}
	e_{j_i}(k)\delta_{j_i}(k) \geq 0,{\;} \forall i = 1..p\label{con_1_2}
	\end{equation}
	\item[ii,]the absolute value of the $i$th elements of \(\boldsymbol{e}_j(k)\)  are no more than \( f_{\phi M}\) times the $i$th  elements of
	\(\boldsymbol{\delta}_j(k)\), $j=1,2...,n$, i.e.
	\begin{equation}
	|{e}_{j_i}(k) | \leq  f_{\phi M}|{\delta}_{j_i}(k) |,{\;} \forall i = 1..p \label{con_2_2}
	\end{equation} 

\end{itemize}

The estimated weights of   sub-systems are updated using \(\boldsymbol{e}_j(k)\) and the output weight matrix $\hat{\mathbf{W}}_n (k+1)$ is updated using \(\boldsymbol{e}_n(k)\) as:
\begin{equation}
\hat{\mathbf{W}}_j(k+1)=\hat{\mathbf{W}}_j(k)+ \alpha_{j} \mathbf{S}_j(k)\hat{\mathbf{W}}_j^{\triangleright T}(k)\mathbf{L}_j(k) \boldsymbol{e}_j(k)  \boldsymbol{x}_j^T(k) \label{eq:updL_2layer_b} 
\end{equation}
\begin{equation}
\hat{\mathbf{W}}_j^{\triangleright}(k+1)=\hat{\mathbf{W}}_j^{\triangleright}(k)+  \alpha_{j}^{\triangleright} \mathbf{L}_j(k) \boldsymbol{e}_j(k)  \hat{\boldsymbol{\sigma}}_{j}^T(k) \label{eq:updL_2layer_a} \\
\end{equation}
\begin{equation}
\hat{\mathbf{W}}_n (k+1)=\hat{\mathbf{W}}_n (k)+ \alpha_{n} \mathbf{L}_n(k) \boldsymbol{e}_n(k)  \boldsymbol{\hat \phi}_{sub}^T(k) \label{eq:updL_2layer_c} \\
\end{equation}
where \( \alpha_{j}^{\triangleright}\), \(\alpha_{j},j=1,..,n-1\) and \(\alpha_{n}\) are constant non-negative scalars,  $\mathbf{ S}_j(k)=diag[\boldsymbol{ \sigma}_{j,1}'(k),...,\boldsymbol{ \sigma}_{j,i}'(k),...,\boldsymbol{ \sigma}_{j,h_j}'(k)] $ are diagonal matrices where    $\boldsymbol{ \sigma}_{j,i}'(k)$ are the gradients of activation functions $\boldsymbol{ \sigma}_{j,i}(k)$, \(\mathbf{L}_j(k) \) and \(\mathbf{L}_n(k) \) are   positive diagonal matrices.   After each update, the weight matrix $\hat{\mathbf{W}}_j$ is shared with the sub-systems from $j+1$ to $n-1$.



To analyze the convergence of the deep FNN, an objective function $V(k)$ of the whole network is proposed as the summation of objective functions of all sub-systems as follows: 

\begin{align}
V(k)&=\resizebox{0.87\linewidth}{!}{$ \sum_{j=1}^{n-1}  \mathbf{Tr}\Big({\Delta \mathbf{W}_{j}^{\triangleright T}(k) \Delta \mathbf{W}_{j}^{\triangleright}(k)}\Big) {+\;}  \sum_{j=1}^{n-1}  \mathbf{Tr}\Big({\Delta \mathbf{W}_{j}^T(k) \Delta \mathbf{W}_{j}(k)}\Big) $}\nonumber\\&{+\;} \resizebox{0.35\linewidth}{!}{$ \mathbf{Tr}\Big( {\Delta \mathbf{W}_{n}^T(k) \Delta \mathbf{W}_{n}(k)}\Big)$}\label{eq:Vk_1}
\end{align}
where $\mathbf{Tr}$ represents the trace of the matrix. 
From \eqref{eq:Vk_1}, for \((k+1)\)th data, the objective function \(V(k+1)\)  can be similarly formulated.
Substituting \eqref{eq:updL_2layer_b},\eqref{eq:updL_2layer_a} and \eqref{eq:updL_2layer_c} into \(V(k+1)\) and taking the difference with \(V(k)\) in \eqref{eq:Vk_1}, it can be shown that : 
\begin{align}
&\Delta  V(k) =V(k+1)-V(k)\nonumber\\
&= \resizebox{0.6\linewidth}{!}{$-2\alpha_{j}^{\vartriangleright}\sum_{j=1}^{n-1} \hat{\boldsymbol{\sigma}}_{j}^T(k)\Delta\mathbf{W}_j^{\triangleright T}(k) \mathbf{L}_j(k)\boldsymbol{e}_j(k) $}
\nonumber\\&
{+\;}\resizebox{0.6\linewidth}{!}{$\boldsymbol{e}_j^T(k) \Big(\sum_{j=1}^{n-1} \alpha_{j}^{{\triangleright}^2}\Vert {\boldsymbol{\hat \sigma}_j}(k) \Vert^2\mathbf{L}_j^T(k)\mathbf{L}_j(k)$} \nonumber\\&  \resizebox{0.92\linewidth}{!}{$+ \sum_{j=1}^{n-1}\alpha_{j}^2\Vert {\boldsymbol{x}_j}(k) \Vert^2 (\mathbf{L}^T_j(k)\hat{\mathbf{W}}_j^{\triangleright}(k)\mathbf{S}^2_j(k)\hat{\mathbf{W}}_j^{\triangleright T}(k)\mathbf{L}_j(k))\Big) \boldsymbol{e}_j(k)$}
\nonumber\\&
{-\;} \resizebox{0.55\linewidth}{!}{$2\alpha_{n} \hat{\boldsymbol{\phi}}_{sub}^T(k)\Delta\mathbf{W}_n^{\triangleright T}(k) \mathbf{L}_n(k)\boldsymbol{e}_n(k)  $}
\nonumber\\&+\resizebox{0.6\linewidth}{!}{$\boldsymbol{e}_n^T(k)  \alpha_{n}^2 \Vert {\boldsymbol{\hat \phi}_{sub}}(k) \Vert^2\mathbf{L}_n^T(k)\mathbf{L}_n(k)\boldsymbol{e}_n(k)  $}
\nonumber\\&{-\;} \resizebox{0.8\linewidth}{!}{$2\alpha_{j} \sum_{j=1}^{n-1} \boldsymbol{e}_j^T(k)(\mathbf{L}^T_j(k)\hat{\mathbf{W}}_j^{\triangleright}(k)\mathbf{S}_j(k)) \Delta\mathbf{W}_j(k)\boldsymbol{x}_j(k)$}
\label{eq:deltaV}\
\end{align}
The training process is divided into 2 phases: a pretraining phase and a fine-tuning phase. The pretrain \cite{nguyen2022analytic} process is conducted by randomly initializing $\mathbf{\hat W}_j(k)=\mathbf{\bar W}_j(k)$ in \eqref{eq:est_yk_2} and \eqref{eq:yk_2}    and then fix  them by setting $\alpha_j=0$ in \eqref{eq:updL_2layer_b}.  Then only train output weight matrix $\mathbf{\hat W}_j^{\vartriangleright}(k)$  and $\mathbf{\hat W}_n(k)$  in initial epochs using update laws \eqref{eq:updL_2layer_a} and \eqref{eq:updL_2layer_c} to reduce errors. 
Since $\mathbf{\hat W}_j(k) $  is fixed as $ \mathbf{\bar W}_j(k)$, then $\boldsymbol{\hat \sigma}_j(k)$ are also fixed as $\boldsymbol{\bar \sigma}_j(k)$.Then there exist ideal weight matrices $\mathbf{W}_j^{\triangleright}$  for each epoch such that equation \eqref{eq:err_2} and \eqref{eq:delta_2} become:
\begin{align}
\boldsymbol{e}_j(k)
=&\boldsymbol{y}(k)-  \boldsymbol{\hat y}_j(k) \\
=&\resizebox{.79\linewidth}{!}{$\boldsymbol{\phi}_{o}\big(\mathbf{W}_j^{\triangleright} \boldsymbol{\bar\sigma}_{j}(\mathbf{\bar W}_j \boldsymbol{x}_j(k))\big)  {-\;}  \boldsymbol{\phi}_{o}\big(\hat{\mathbf{W}}_j^{\triangleright}(k) \boldsymbol{\bar\sigma}_{j}(\bar{\mathbf{W}}_j(k) \boldsymbol{x}_j(k))\big) $} \nonumber\label{eq:err_2fix}
\end{align}
\begin{equation}
\boldsymbol{\delta}_j(k) = \mathbf{W}_j^{\triangleright} \boldsymbol{\bar \sigma}_{j}(k) - \hat{\mathbf{W}}_j^{\triangleright}(k) {\boldsymbol{\bar \sigma}}_{j}(k)=\Delta{\mathbf{W}}_j^{\triangleright}(k) {\boldsymbol{\bar \sigma}}_{j}(k)\label{eq:delta_2fix}
\end{equation}  
where $\boldsymbol{\hat \sigma}_{j}(k)=\boldsymbol{\bar \sigma}_{j}(k)$ in     \eqref{eq:deltaV}    in pretraining. Substitute \eqref{eq:delta_2fix} and \eqref{eq:deltak_n} into  \eqref{eq:deltaV}, with $\alpha_j=0$, the change in objective function for pretraining reduces to  : 
\begin{align}
&\Delta V_{pre}(k) 
= \resizebox{.79\linewidth}{!}{${-}2\alpha_{j}^{\vartriangleright}\sum_{j=1}^{n-1} \boldsymbol{\delta}_j^T(k)\mathbf{L}_j(k)\boldsymbol{e}_j(k)-2\alpha_{n} \boldsymbol{\delta}_n^T(k)\mathbf{L}_n(k)\boldsymbol{e}_n(k)$}
 \nonumber\\ 
&{+\;}\resizebox{0.7\linewidth}{!}{$\boldsymbol{e}_j^T(k) \sum_{j=1}^{n-1} \alpha_{j}^{{\triangleright}^2 }\Vert {\boldsymbol{\bar \sigma}_j}(k) \Vert^2\mathbf{L}_j^T(k)\mathbf{L}_j(k)   \boldsymbol{e}_j(k) $}
\nonumber\\&+\resizebox{0.65\linewidth}{!}{$\boldsymbol{e}_n^T(k)  \alpha_{n}^2 \Vert {\boldsymbol{\hat \phi}_{sub}}(k) \Vert^2\mathbf{L}_n^T(k)\mathbf{L}_n(k)\boldsymbol{e}_n(k) $}
\label{eq:deltaV_pre}
\end{align}
\noindent 
Let $L_{jM}$, $L_{nM}$ be the maximum eigenvalues of $\boldsymbol L_j(k),\boldsymbol L_n(k)$. Since $ {\boldsymbol{\bar \sigma}_j}(k)$ and $ {\boldsymbol{\hat \phi}_{sub}}(k) $ are bounded, there exist    positive constants \(d^{pre}_j \) and \(d^{pre}_n \)  such that: 
\begin{align}
	&\resizebox{0.7\linewidth}{!}{$\boldsymbol{e}_j^T(k)  \sum_{j=1}^{n-1} \alpha_{j}^{{\triangleright}^2 }\Vert {\boldsymbol{\bar \sigma}_j}(k) \Vert^2\mathbf{L}_j^T(k)\mathbf{L}_j(k) \boldsymbol{e}_j(k) $} \nonumber
\\&\resizebox{0.7\linewidth}{!}{$+\boldsymbol{e}_n^T(k)  \alpha_{n}^2 \Vert {\boldsymbol{\hat \phi}_{sub}}(k) \Vert^2\mathbf{L}_n^T(k)\mathbf{L}_n(k)\boldsymbol{e}_n(k)$} \label{pp}\\& \resizebox{0.85\linewidth}{!}{$\leq \sum_{j=1}^{n-1}\alpha_{j}^{{\triangleright}^2}d^{pre}_j L_{jM}^2   \Vert {\boldsymbol{e}_j}(k) \Vert^2 +\alpha_{n}^{2}d^{pre}_n  L_{nM}^2   \Vert {\boldsymbol{e}_n}(k) \Vert^2 $}\nonumber
\end{align}
\textcolor{black}{where $d_j^{pre} $ is the norm bound for any $\Vert\ {\boldsymbol{\bar \sigma}_j}(k)\Vert^2$ and $d_n^{pre} $ is the norm bound for any $ \Vert {\boldsymbol{\hat \phi}_{sub}}(k) \Vert^2$.}
Substituting inequality  \eqref{pp} into the second and third row of \eqref{eq:deltaV_pre} and using inequalities in \eqref{con_1_2}, \eqref{con_2_2} in the first row  of \eqref{eq:deltaV_pre}  , we have: 
\begin{align}
&\resizebox{1\linewidth}{!}{$\Delta V_{pre}(k) 
\leq-2\alpha_j^{\vartriangleright}\sum_{j=1}^{n-1} \boldsymbol{\delta}_j^T(k)\mathbf{L}_j(k)\boldsymbol{e}_j(k) {+\;}\sum_{j=1}^{n-1}\alpha_{j}^{{\triangleright}^2}d^{pre}_j  L_{jM}^2   \Vert {\boldsymbol{e}_j}(k) \Vert^2$}
\nonumber\\& 
\resizebox{0.7\linewidth}{!}{$ 
-2\alpha_n  \boldsymbol{\delta}_n^T(k)\mathbf{L}_n(k)\boldsymbol{e}_n(k) {+\;} \alpha_{n}^{ 2}d^{pre}_n  L_{nM}^2   \Vert {\boldsymbol{e}_n}(k) \Vert^2$}
\nonumber\\& 
\resizebox{0.7\linewidth}{!}{$ \leq -\sum_{j=1}^{n-1}\alpha_j^{\vartriangleright}(\frac{2L_{jm}}{f_{\phi M}}-\alpha_{j}^{{\triangleright} }d^{pre}_j  L_{jM}^2) \Vert {\boldsymbol{e}_j}(k)\Vert^2 $}\nonumber\\&-\resizebox{0.6\linewidth}{!}{$  \alpha_n(\frac{2L_{nm}}{f_{\phi M}}-\alpha_nd^{pre}_n  L_{nM}^2) \Vert {\boldsymbol{e}_n}(k)\Vert^2
\label{ieq:originalpre} $}
\end{align}
where   $L_{jm},L_{nm}$ are the minimum eigenvalues of   matrix $\boldsymbol L_j(k),\boldsymbol L_n(k)$. It can be shown that if $\frac{2L_{jm}}{f_{\phi M}}-\alpha_j^{\vartriangleright}d^{pre}_j  L_{jM}^2 > 0
$ and $\frac{2L_{nm}}{f_{\phi M}}-\alpha_nd^{pre}_n  L_{nM}^2 > 0
$ then \(\Delta V_{pre}(k)\leq0\), which guarantees the convergence in pretraining.
\textcolor{black}{
\begin{theorem}
Consider the deep collaborative fully connected neural network given by (6) with the update laws (14), (15) and (16) in the fine-tuning phase.  Let the non-negative scalar $\alpha_j$ in (14) be chosen as ${\alpha_j=\alpha}_j^\vartriangleright$ and $\mathbf L_j\left(k\right)$ and $\mathbf{L}_{n}\left(k\right)$ in (14), (15) and (16) be chosen to satisfy the following conditions:
\begin{equation}\label{condition}
\frac{2L_{jm}}{f_{\phi M}}-\alpha_j^\vartriangleright d_jL_{jM}^2>0,\frac{2L_{nm}}{f_{\phi M}}-\alpha_nd_nL_{nM}^2>0
\end{equation}
where $L_{jM}$ and $L_{nM}$ are the maximum eigenvalues of $\mathbf{L}_{j}\left(k\right)$ and $\mathbf{L}_{n}\left(k\right)$, $d_j$ and $d_n$ are positive scalars.  The output training errors $\boldsymbol{e}_{j}\left(k\right)$ and $\boldsymbol{e}_{n}\left(k\right)$ converge towards 0 as $k\rightarrow\infty$.
\end{theorem}}

\begin{remark}
After pretraining, update law \eqref{eq:updL_2layer_b} is activated by setting $\alpha_j$ to   equal to non-zero constants $\alpha_j^{\vartriangleright}$. In this fine-tuning process, all layers' weights are updated together to further reduce output error and improve performance. 
From \eqref{eq:deltak_2} and \eqref{eq:deltak_n}, we have:
\begin{equation}
\begin{aligned}
\resizebox{.87\hsize}{!}{$\Delta\mathbf{W}_j^{\triangleright}(k)\hat{\boldsymbol{\sigma}}_{j}(k) =\boldsymbol{\delta}_j(k)-\hat{\mathbf{W}}_j^{\triangleright}(k)\Delta \boldsymbol{\sigma}_{j}(k) -\Delta\mathbf{W}_j^{\triangleright}(k) \Delta \boldsymbol{\sigma}_{j}(k)$}
\label{deltaj}
\end{aligned}
\end{equation}
\begin{equation}
\begin{aligned}
\Delta\mathbf{W}_n^{\triangleright}(k)\hat{\boldsymbol{\phi}}_{sub}(k) 
&=\boldsymbol{\delta}_n(k)\label{deltan}
\end{aligned}
\end{equation}
Substituting  \eqref{deltaj} and  \eqref{deltan} into \eqref{eq:deltaV}, we have:
\begin{align}
&\Delta V(k) 
= \resizebox{0.45\linewidth}{!}{${-\;}2\alpha_{j}^{\vartriangleright}\sum_{j=1}^{n-1} \boldsymbol{\delta}_j^T(k)\mathbf{L}_j(k)\boldsymbol{e}_j(k)$}
 \nonumber\\ 
&\resizebox{0.6\linewidth}{!}{${+\;}\boldsymbol{e}_j^T(k) \Big(\sum_{j=1}^{n-1} \alpha_{j}^{{\triangleright}^2} \Vert {\boldsymbol{\hat \sigma}_j}(k) \Vert^2\mathbf{L}_j^T(k)\mathbf{L}_j(k)$}   \nonumber\\&\resizebox{0.95\linewidth}{!}{$+ \sum_{j=1}^{n-1}{\alpha_{j}^{\vartriangleright}}^2\Vert {\boldsymbol{x}_j}(k) \Vert^2 (\mathbf{L}^T_j(k)\hat{\mathbf{W}}_j^{\triangleright}(k)\mathbf{S}^2_j(k)\hat{\mathbf{W}}_j^{\triangleright T}(k)\mathbf{L}_j(k))\Big) \boldsymbol{e}_j(k)$} \nonumber\\&\resizebox{0.4\linewidth}{!}{$-2\alpha_{n}\boldsymbol{\delta}_n^T(k)\mathbf{L}_n(k)\boldsymbol{e}_n(k)$}
\nonumber\\&\resizebox{0.6\linewidth}{!}{$+\boldsymbol{e}_n^T(k)  \alpha_{n}^2 \Vert {\boldsymbol{\hat \phi}_{sub}}(k) \Vert^2\mathbf{L}_n^T(k)\mathbf{L}_n(k)\boldsymbol{e}_n(k) $}
\nonumber\\
&\resizebox{1\linewidth}{!}{${+\;}2\alpha_{j}^{\vartriangleright}\sum_{j=1}^{n-1}\boldsymbol{e}_j^T(k)\mathbf{L}_j^T(k) \hat{\mathbf{W}}_j^{\triangleright}(k)\Big( \Delta{\boldsymbol{\sigma}_{j}}(k)  {-\;}  \mathbf{S}_j(k) \Delta\mathbf{W}_j(k)\boldsymbol{x}_j(k)   \Big)$}
\nonumber\\
&\resizebox{0.65\linewidth}{!}{${+\;}\sum_{j=1}^{n-1}2\alpha_{j}^{\vartriangleright}\Delta \boldsymbol{\sigma}_{j}^T(k) \Delta\mathbf{W}_j^{\triangleright T}(k)\mathbf{L}_j(k)\boldsymbol{e}_j(k) $}
\label{eq:deltaV_3}
\end{align}
Consider the terms in the second  last row in \eqref{eq:deltaV_3}, using  Taylor expansion   we have:
\begin{equation}
\begin{aligned}
\Delta&{\boldsymbol{\sigma}_{j}}(k)=\boldsymbol{\sigma}_j(\mathbf{W}_j\boldsymbol{x}_j(k))-\boldsymbol{\sigma}_j(\mathbf{\hat W}_j(k)\boldsymbol{x}_j(k))\\&=\mathbf{S}_j(k)\Delta{\mathbf{W}}_j(k)\boldsymbol{x}_j(k)+\mathbf{O}^{t}\big(\Delta{\mathbf{W}}_j(k)\boldsymbol{x}_j(k)\big)
\label{taylor}
\end{aligned}
\end{equation}

where  $\mathbf{ S}_j(k)  $ are diagonal matrices   with elements of gradients of activation functions $\boldsymbol{ \sigma}_{j,i}(k)$, $\mathbf{O}^{t}\big(\Delta{\mathbf{W}}_j(k)\boldsymbol{x}_j(k)\big)$ are summation of high order terms. In the fine-tuning phase after pre-training phase, the errors are sufficiently small and hence the higher order terms in last 2 terms of \eqref{eq:deltaV_3}, which are $O^3$ and more are negligible as compared to the other terms which are of $O^2$.

Then  equation \eqref{eq:deltaV_3} becomes:
\begin{equation}
\begin{aligned}
&\Delta V(k) = \resizebox{0.8\linewidth}{!}{$-{\;}2\alpha_j^{\vartriangleright}\sum_{j=1}^{n-1}\boldsymbol{\delta}_j^T(k)\mathbf{L}_j(k)\boldsymbol{e}_j(k) -2\alpha_n\boldsymbol{\delta}_n^T(k)\mathbf{L}_n(k)\boldsymbol{e}_n(k)$}\\
&\resizebox{0.7\linewidth}{!}{${+\;}\boldsymbol{e}_j^T(k) \Big(\sum_{j=1}^{n-1} \alpha_{j}^{{\triangleright}^2} \Vert {\boldsymbol{\hat \sigma}_j}(k) \Vert^2\mathbf{L}_j^T(k)\mathbf{L}_j(k) $} \\&\resizebox{1\linewidth}{!}{$+ \sum_{j=1}^{n-1}{\alpha_{j}^{\vartriangleright}}^2\Vert {\boldsymbol{x}_j}(k) \Vert^2 (\mathbf{L}^T_j(k)\hat{\mathbf{W}}_j^{\triangleright}(k)\mathbf{S}^2_j(k)\hat{\mathbf{W}}_j^{\triangleright T}(k)\mathbf{L}_j(k))\Big) \boldsymbol{e}_j(k)$}
\\& +\boldsymbol{e}_n^T(k)  \alpha_{n}^2 \Vert {\boldsymbol{\hat \phi}_{sub}}(k) \Vert^2\mathbf{L}_n^T(k)\mathbf{L}_n(k)\boldsymbol{e}_n(k)  
\label{eq:deltaV_5}
\end{aligned}
\end{equation}

\textcolor{black}{Since the smooth activation functions $\boldsymbol{  \sigma}$ and ${\boldsymbol{ \phi}_{sub}}$ are saturated, therefore ${\boldsymbol{\hat \sigma}_j}(k)$, ${\boldsymbol{\hat \phi}_{sub}}(k)$ and  ${\boldsymbol{ x}_{j}(k)}$ are bounded. Since the initial weights matrices $\hat{\mathbf{W}}_j^{\triangleright}(0)$, $\hat{\mathbf{W}}_j(0)$ and $\hat{\mathbf{W}}_n(0)$ are bounded, so according to update laws \eqref{eq:updL_2layer_b}, \eqref{eq:updL_2layer_a} and \eqref{eq:updL_2layer_c}(refer to Appendix),  there exist positive constants \(d_j ,d_n  \)   such that :
	\begin{flalign}\label{djdn}
	&\boldsymbol{e}_j^T(k) \Big(\sum_{j=1}^{n-1} \alpha_{j}^{{\triangleright}^2 }\Vert {\boldsymbol{\hat \sigma}_j}(k) \Vert^2\mathbf{L}_j^T(k)\mathbf{L}_j(k)  \\
	&\resizebox{1\linewidth}{!}{$+ \sum_{j=1}^{n-1}{\alpha_{j}^{\vartriangleright}}^2\Vert {\boldsymbol{x}_j}(k) \Vert^2 (\mathbf{L}^T_j(k)\hat{\mathbf{W}}_j^{\triangleright}(k)\mathbf{S}^2_j(k)\hat{\mathbf{W}}_j^{\triangleright T}(k)\mathbf{L}_j(k))\Big) \boldsymbol{e}_j(k)$} \nonumber
\\&+\boldsymbol{e}_n^T(k)  \alpha_{n}^2 \Vert {\boldsymbol{\hat \phi}_{sub}}(k) \Vert^2\mathbf{L}_n^T(k)\mathbf{L}_n(k)\boldsymbol{e}_n(k) \nonumber\\& \resizebox{0.9\linewidth}{!}{$\leq \sum_{j=1}^{n-1} \alpha_j^{{\vartriangleright}^2}d_j L_{jM}^2   \Vert {\boldsymbol{e}_j}(k) \Vert^2  + \alpha_n^2 d_n  L_{nM}^2   \Vert {\boldsymbol{e}_n}(k) \Vert^2$}\nonumber
	\end{flalign} 
where $d_j $ is the norm bound for any \resizebox{0.4\linewidth}{!}{$\Vert{\boldsymbol{x}_j}(k)\Vert^2\Vert\hat{\mathbf{W}}_j^{\triangleright}(k)\mathbf{S}_j(k) \Vert^2$} and $d_n $ is the norm bound for any $ \Vert{\boldsymbol{\hat\phi}_{sub}}(k)\Vert^2$.}

Therefore, \eqref{eq:deltaV_5} becomes:
\begin{align}
&\resizebox{1\linewidth}{!}{$\Delta V (k) 
\leq-2\alpha_j^{\vartriangleright}\sum_{j=1}^{n-1} \boldsymbol{\delta}_j^T(k)\mathbf{L}_j(k)\boldsymbol{e}_j(k) {+\;}\sum_{j=1}^{n-1}\alpha_j^{{\vartriangleright}^2}d_j   L_{jM}^2   \Vert {\boldsymbol{e}_j}(k) \Vert^2$}
\nonumber\\& 
\resizebox{0.7\linewidth}{!}{$ 
-2\alpha_n  \boldsymbol{\delta}_n^T(k)\mathbf{L}_n(k)\boldsymbol{e}_n(k) {+\;} \alpha_n^{ 2}d_n   L_{nM}^2   \Vert {\boldsymbol{e}_n}(k) \Vert^2$}
\label{ieq:original} 
\end{align}
Using inequalities in \eqref{con_1_2}, \eqref{con_2_2} in the first term of (\ref{ieq:original}), we have:
\begin{equation}
\begin{aligned}
-2 \boldsymbol{\delta}^T_j(k)\mathbf{L}_j(k)\boldsymbol{e}_j(k)\leq \frac{-2\alpha_j^{\vartriangleright}L_{jm}}{f_{\phi M}} \Vert {\boldsymbol{e}_j}(k)\Vert^2 
\end{aligned}
\end{equation}
\begin{equation}
\begin{aligned}
-2 \boldsymbol{\delta}^T_n (k)\mathbf{L}_n (k)\boldsymbol{e}_n (k)\leq \frac{-2\alpha_n L_{n m}}{f_{\phi M}} \Vert {\boldsymbol{e}_n }(k)\Vert^2
\end{aligned}
\end{equation}
Rearranging the equation \eqref{ieq:original}, we can get:
\begin{align}\label{Vfinalcondition}
&\Delta V(k) \leq -\sum_{j=1}^{n-1}\alpha_j^{\vartriangleright}(\frac{2L_{jm}}{f_{\phi M}}
-\alpha_j^{\vartriangleright} d_j L_{jM}^2) \Vert {\boldsymbol{e}_j}(k)\Vert^2\nonumber\\&- \alpha_n(\frac{2L_{nm}}{f_{\phi M}}-\alpha_n d_n L_{nM}^2) \Vert {\boldsymbol{e}_n}(k)\Vert^2
\end{align}

\textcolor{black}{as seen from equation \eqref{Vfinalcondition}, when the conditions in \eqref{condition} are satisfied, then the two terms on the right-hand side of  \eqref{Vfinalcondition} are negative definite in $\boldsymbol{e}_{j}\left(k\right)$ and $\boldsymbol{e}_{n}\left(k\right)$,  and hence \(\Delta V(k)\leq0\).} Since  \( V(k)\) is  non-negative and bounded from below, then it is converging as $k$ increases, which also indicates that all errors are converging for each epoch. This condition in \eqref{condition} adjusts the gain matrix \(\mathbf{L}_j(k)\) after  it is randomly initialized in each update. \textcolor{black}{The proof is complete.}
	\end{remark}

\section{Case Studies}
In this section, a case study using MNIST dataset  was first done to show the performance of the approximation ability in classification tasks. Then an online robot kinematics control task was done using an industrial robot UR5e to illustrate the regression approximation ability and online adaptation ability of the proposed algorithm.
\subsection{MNIST Dataset Case Studies}
Case Studies were  performed on a fully connected network based on handwritten number dataset MNIST.
 MNIST dataset is a 10 classes image dataset with the input image size of $28*28$. The case study  is done on a  fully connected neural network  with 3 subsystems (subsystem I:784-Sig-150-Sig-10-Sig, subsystem II:784-Sig-150-Sig-100-Sig-10-Sig, subsystem III:784-Sig-150-Sig-100-Sig-50-Sig-10-Sig) connected by a collaborative layer (see Fig \ref{fig:endtoend} also):

 Unlike SGD, where learning rate can be selected from several empirical trials, the choice of $\mathbf L_j(k)$ in proposed algorithm in each step can be  automatically reduces  by  the condition given in \eqref{condition} and convergence can always be assured. The pretraining was first conducted by setting $\alpha_j=0$ in \eqref{eq:updL_2layer_b} and $\mathbf{L}_j=\mathbf{L}_n=diag(0.01,...,0.01)$ for 10 epochs and then by setting $\alpha_j^{\vartriangleright}=\alpha_j=\alpha_n=1$ the fine-tuning process was conducted.
 To obtain the optimal test performance for both methods, we have tuned the hyperparameters of both SGD and the proposed method and compared their results.


The training and testing accuracies are shown in Table \ref{tabel1} compared between SGD and the proposed method. The sub-system column refers to each sub-system  performance  as shown in Fig \ref{fig:endtoend}(a)-(d). It can be observed from the last row of Table \ref{tabel1} that the four hidden layer fully connected neural network trained with SGD has already dropped to 89.3 percent due to the gradient vanishing problem of using Sigmoid activation. However, our proposed learning algorithm allows the use of  Sigmoid activation   for deeper neural networks without gradient vanishing problem. The results also show that by using collaborative learning, the testing accuracy has a noticeable improvement. To further illustrate the problem of gradient vanishing using Sigmoid activation functions, we tested the proposed method and SGD with deeper networks (subsystem IV:784-Sig-150-Sig-100-Sig-50-Sig-50-Sig-50-Sig-10-Sig, subsystem V:784-Sig-150-Sig-100-Sig-50-Sig-50-Sig-50-Sig-50-Sig-10-Sig) as stated in the fourth and fifth row  of Table \ref{tabel1}. It was observed  that SGD could not converge with deeper layers, but the collaborative network with deeper subsystems included could converge and   achieve a final accuracy of  $98.6\%$.

It is shown that compared with SGD, our result on the MNIST can achieve similar performance. From Table \ref{tabel1}, with collaborative learning, the accuracy has a noticeable improvement. This demonstrates the potential of using smooth activation functions in deep control systems.

\begin{table}
\renewcommand{\arraystretch}{0.4} 
\caption{\textcolor{black}{Test results on Collaborative learning of  Fully connected networks with MNIST}}
\centering
\begin{tblr}{
  cells = {c},
  cell{1}{1} = {r=2}{},
  cell{1}{2} = {c=2}{},
  cell{1}{4} = {c=2}{},
  cell{1}{6} = {c=2}{},
  cell{3}{6} = {r=3}{},
  cell{3}{7} = {r=3}{},
  vlines,
  hline{1,3,6,8} = {-}{},
  hline{2} = {2-7}{},
  hline{4-7} = {1-5}{},
}
Network                                     & SGD              &                 & sub-system       &                 & Collaborative    &                 \\
                                            & {Train\\ acc } & {Test\\ acc } & {Train\\ acc } & {Test\\ acc } & {Train\\ acc } & {Test\\ acc } \\
{sub-system I}                  & 99.5             & 98.2            & 99.3             & 98.2            & 99.8             & 98.7            \\
{sub-system III}          & 99.9             & 98.3            & 99.6             & 98.4            &                  &                 \\
{sub-system II} & 99.8             & 89.3            & 99.5             & 98.3            &                  &             
 \\
{sub-system IV} & {Diverge} & {Diverge} & 99.2          & 98.1    &99.85& 98.6    \\  
{sub-system V}    & {Diverge}  & {Diverge}  & 99.0          & 98.0                
\end{tblr}
\label{tabel1}
\end{table}

\subsection{Online Jacobian matrix approximation task on UR5E}

Jacobian matrix, mapping from joint space to Cartesian space, directly gives feedback from Cartesian space and therefore is crucial to real-time control\cite{cheah2015task}. In this section,  the Jacobian matrix of UR5E robot arm with unknown kinematics is approximated by a  FNN using the proposed   algorithm.

Let  $k$ represent the sampling time, the  Cartesian space end effector velocities $\boldsymbol{\dot x}$  and joint velocities  $\boldsymbol{\dot q}$ has the following relationship:
\begin{equation}
\dot{\boldsymbol{x}}(k)= \mathbf{J}(\boldsymbol{q}(k))\dot{\boldsymbol{q}}(k) \label{eq:jacob_all}
\end{equation}
where $\mathbf{J}(\boldsymbol{q}(k))$ is the  Jacobian matrix.

Jacobian matrix is approximated by deep FNNs using the proposed algorithm. The estimated Jacobian matrices $\hat{\mathbf{J}}_j  $ are retrieved  from the $j$th, $j=1,...,n-1$ sub-system  as \eqref{eq:est_yk_2} and the  velocity in Cartesian space is approximated by the collaborated network as \eqref{eq:est_yk_n} :
\begin{equation}
\begin{aligned} 
\hat{\dot{\boldsymbol{ x}}}_j(k)&=\resizebox{0.4\linewidth}{!}{$\hat{\mathbf{J}}_j(\boldsymbol{q}(k),\mathbf{\hat W}_j^\vartriangleright,\mathbf{\hat W}_j)\dot{\boldsymbol{q}}(k)$}
\\&= \resizebox{0.55\linewidth}{!}{$\sum_{h=1}^{r}{\boldsymbol \phi}_{o_h}\big(\mathbf{\hat W}_j^\vartriangleright(k)\boldsymbol{\hat\sigma}_{j}(k)\big){\dot q}_m(k)$}
\label{eq:est_kxj}
\end{aligned}
\end{equation}
\begin{equation}
\begin{aligned} 
&\hat{\dot{\boldsymbol{ x}_n}}(k)
= \resizebox{0.6\linewidth}{!}{$ \sum_{h=1}^{r}\boldsymbol \phi_{{c}_h}\big(\mathbf{\hat W}_n(k) \boldsymbol{\hat \phi}_{sub}(k)\big){\dot q}_m(k))$}
\label{eq:est_kxn}
\end{aligned}
\end{equation}
where   \(\boldsymbol{\hat\sigma}_{j }(k)=\boldsymbol{\sigma}_{j }(\mathbf{ \hat W}_j(k)\boldsymbol{ q}_j(k)))\), \(\boldsymbol{\hat \phi}_{sub}(k)= [\boldsymbol{\hat  {\dot x}}_1(k),\boldsymbol{\hat   {\dot x}}_2(k),..,\boldsymbol{\hat  {\dot x}}_j(k),..,\boldsymbol{\hat  {\dot x}}_n(k)]^T\),  $r$ denotes the number of joints, $\boldsymbol{ q}_j(k)$ is the input joint angle of the $j$th sub-system, ${\dot q}_m(k)$ is the $m$th element of joint velocity $\boldsymbol{\dot q}(k)$,  the activation functions ${\boldsymbol\phi}_c$ and  ${\boldsymbol \phi}_o$  are chosen as linear  activation function in the following online kinematics task.

 

The output error $\boldsymbol{e}_j(k)$, which is formulated using  \eqref{eq:est_kxj}, is utilized to update the weights of the $j$th sub-system. The collaborative layer is updated based on $\boldsymbol{e}_n(k)$ formulated using \eqref{eq:est_kxn}: 
\begin{flalign}
&\boldsymbol{e}_j(k) = {\boldsymbol{\dot x}}(k)-\hat{\dot{\boldsymbol{ x}}}_j(k)\\& \resizebox{1\linewidth}{!}{$ =\sum_{h=1}^{r}{\boldsymbol \phi}_{o_h}\big(\mathbf{W}_j^{\vartriangleright}(k)\boldsymbol{\sigma}_{j-1}(k)\big){\dot q}_m(k))
{-}\sum_{h=1}^{r}{\boldsymbol \phi}_{o_h}\big(\mathbf{\hat W}_j^\vartriangleright(k)\boldsymbol{\hat\sigma}_{j-1}(k)\big){\dot q}_m(k)$}\label{eq:dyna_errj}\nonumber
\end{flalign}
\begin{flalign}
&\boldsymbol{e}_n(k) =  {\boldsymbol{\dot x}}(k)-\hat{\dot{\boldsymbol{ x}_n}}(k)\\&\resizebox{1\linewidth}{!}{$= \sum_{h=1}^{r}\boldsymbol \phi_{{c}_h}\big(\mathbf{W}_n(k) \boldsymbol{\hat \phi}_{sub}(k)\big){\dot q}_m(k)) 
{-} \sum_{h=1}^{r}\boldsymbol \phi_{{c}_h}\big(\mathbf{\hat W}_n(k)\boldsymbol{\hat \phi}_{sub}(k)\big){\dot q}_m(k))$}\nonumber\label{eq:dyna_errn}
\end{flalign}
where $
\dot{\boldsymbol{x}}(k)
= \sum_{h=1}^{r}{\boldsymbol \phi}_{o_h}\big(\mathbf{W}_j^{\vartriangleright}(k)\boldsymbol{\sigma}_{j }(k)\big){\dot q}_m(k)) $
and
$ 
\dot{\boldsymbol{x}}(k)
= \sum_{h=1}^{r}\boldsymbol \phi_{{c}_h}\big(\mathbf{W}_n(k) \boldsymbol{\hat \phi}_{sub}(k)\big){\dot q}_m(k)
$,  \( \boldsymbol{\sigma}_{j }(k)=\boldsymbol{\sigma}_{j }(\mathbf{  W}_j(k)\boldsymbol{ q}_j(k))) \). Comparing with  \eqref{eq:err_2} and \eqref{eq:err_nn}, the error \(\boldsymbol{e}_j(k)\) and \(\boldsymbol{e}_n(k)\) are identical to  the output error of the sub-systems.  Therefore, the  errors \(\boldsymbol{e}_j(k) \) and \(\boldsymbol{e}_n(k)\) are updated following  the update laws in \eqref{eq:updL_2layer_b}, \eqref{eq:updL_2layer_a} and \eqref{eq:updL_2layer_c} with guaranteed convergence.

Then, after all the subsystems have been trained,  
The Jacobian matrix \(\hat{\mathbf{J}}_{n-1}(\boldsymbol{q}(k),\mathbf{\hat W}_ {n-1}^\vartriangleright,\mathbf{\hat W}_ {n-1})\)  of the $n-1$th sub-system  alone is applied in the   online kinematic control to 
accommodate changes. The reference joint velocity  \(\dot{\boldsymbol{q}}\) for online control is defined as :
\begin{equation}
\begin{aligned}
\resizebox{0.85\linewidth}{!}{$\dot{\boldsymbol{q}}(k)  = \hat{\mathbf{J}}_{n-1}^\dagger(\boldsymbol{q}(k),\mathbf{\hat W}_ {n-1}^\vartriangleright,\mathbf{\hat W}_ {n-1})(\dot{\boldsymbol{x}}_d(k) -\alpha\Delta\boldsymbol{x}(k)) $}
%
\label{eq:control_qre}
\end{aligned}
\end{equation}
 where $\alpha\Delta \boldsymbol{x}(k)=\boldsymbol{x}(k)-\boldsymbol{x}_d(k),\alpha$ is a positive scalar,   \({\boldsymbol{x}}_d(k),\dot{\boldsymbol{x}}_d(k)\) represents the desired position and velocity of the end effector in the  task space \(\hat{\mathbf{J}}_{n-1}^\dagger\) is the pseudoinverse matrix of  \(\hat{\mathbf{J}}_{n-1}\).
By multiplying \eqref{eq:control_qre} with \(\hat{\mathbf{J}}_{n-1}(\boldsymbol{q}(k),\mathbf{\hat W}_ {n-1}^\vartriangleright,\mathbf{\hat W}_ {n-1})\), we have:
\begin{eqnarray}
\hat{\mathbf{J}}_{n-1}((\boldsymbol{q}(k),\mathbf{\hat W}_ {n-1}^\vartriangleright,\mathbf{\hat W}_ {n-1})\dot{\boldsymbol{q}}(k) = \dot{\boldsymbol{x}}_d(k) -\alpha\Delta \boldsymbol{x}(k) \label{eq:premul_qre}
\end{eqnarray}
Subtracting  \eqref{eq:jacob_all} and \eqref{eq:premul_qre} we have:
\begin{equation}
\begin{aligned}
&{\mathbf{J}}(\boldsymbol{q}(k))\dot{\boldsymbol{q}}(k)-\hat{\mathbf{J}}_{n-1}(\boldsymbol{q}(k),\mathbf{\hat W}_ {n-1}^\vartriangleright,\mathbf{\hat W}_ {n-1})\dot{\boldsymbol{q}}(k) 
\\
=& \dot{{\boldsymbol{x}}}(k)-\dot{\boldsymbol{x}}_d(k) +\alpha\Delta \boldsymbol{x}(k)=\Delta \dot{\boldsymbol{x}}(k) +\alpha\Delta \boldsymbol{x}(k) \label{eq:dyna_erroo}
\end{aligned}
\end{equation}
During online learning, online feedback error is constructed as \(\boldsymbol{\varepsilon}(k) = \Delta \dot{\boldsymbol{x}}(k) +\alpha\Delta \boldsymbol{x}(k)\), from \eqref{eq:jacob_all},\eqref{eq:est_kxj}, we have:
\begin{equation}
\begin{aligned}
\resizebox{.85\linewidth}{!}{$\boldsymbol{\varepsilon}(k) = {\mathbf{J}}(\boldsymbol{q}(k))\dot{\boldsymbol{q}}(k)-\hat{\mathbf{J}}_{n-1}(\boldsymbol{q}(k),\mathbf{\hat W}_ {n-1}^\vartriangleright,\mathbf{\hat W}_ {n-1} )\dot{\boldsymbol{q}}(k)$}
 \label{eq:dyna_erre}
\end{aligned}
\end{equation}

Hence, in the online kinematics control, update laws \eqref{eq:updL_2layer_b},\eqref{eq:updL_2layer_a}   based on  $\boldsymbol{\varepsilon}(k)$ are used, which ensures  the convergence of the online feedback error.

A trajectory tracking control experiment of the robot arm in task space was conducted to show the proposed algorithm's approximation ability in  online kinematics control tasks.  For approximating the Jacobian matrix in the online task, a deep collaborated FNN  was employed and evaluated  with following the structure with three subsystems (subsystem I:3-Sig-12-Sig-3-Sig, subsystem II:3-Sig-12-Sig-24-Sig-3-Sig, subsystem III:3-Sig-12-Sig-24-Sig-24-Sig-3-Sig) connected by a collabrative layer(see Fig \ref{fig:endtoend} also):

 The input data of the network is $[\boldsymbol{ q},\boldsymbol{\dot q}]$ and the output is end-effector velocity $\boldsymbol{\dot x}$. All the data are acquired from the robot arm real-time data exchange every 0.01s. In the FNN, Sigmoid activation functions were employed for the hidden layers in each subsystem and identity activation functions were chosen for  each output layer. 

\begin{figure}[!h]
	\centering
	\begin{minipage}[t]{ \linewidth}
		\centering
		\subfigure[using offline learning]{
		\includegraphics[width=\linewidth]{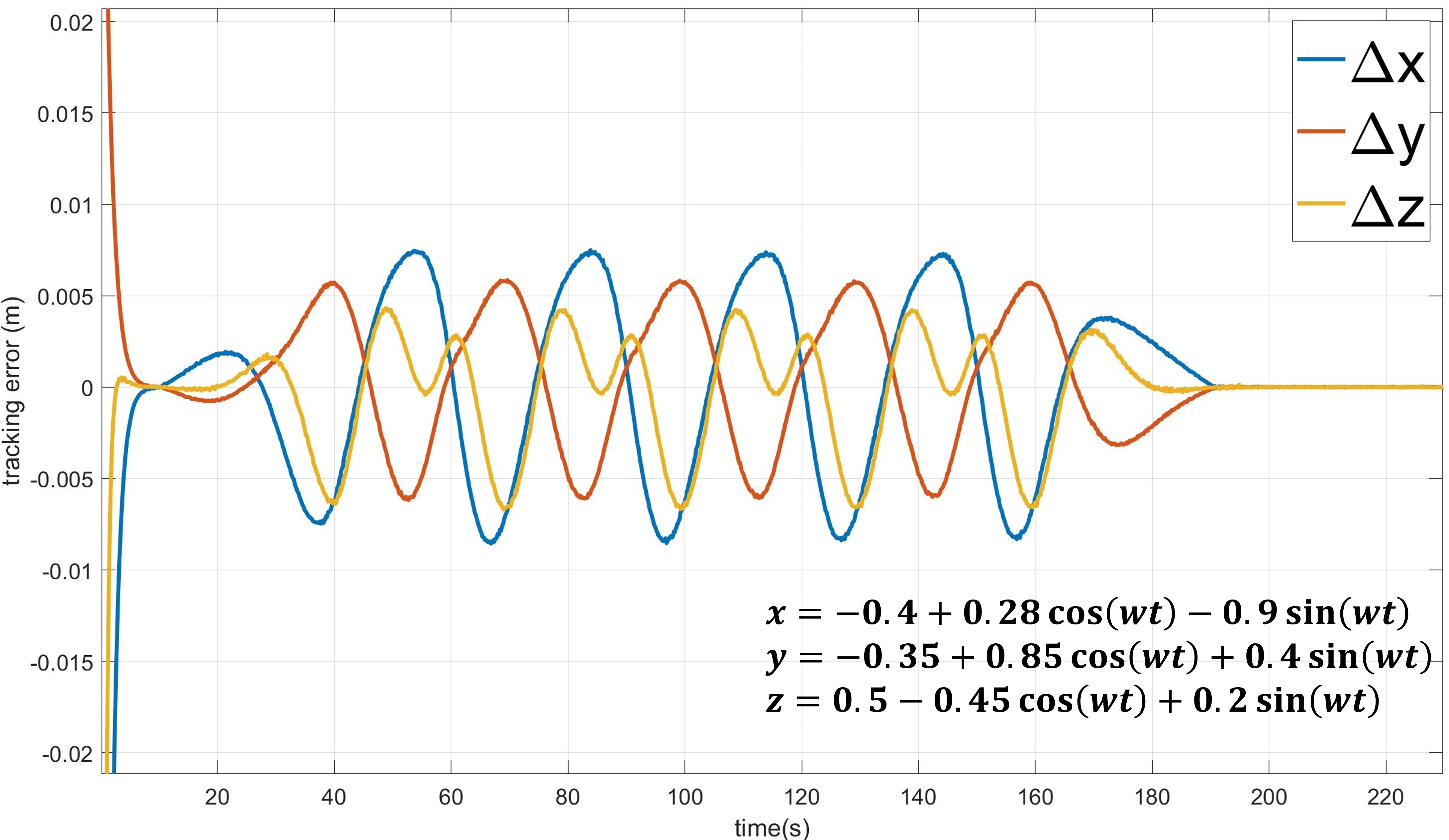} \label{fig:noonline}}
	\end{minipage}
	\begin{minipage}[t]{ \linewidth}
		\centering
		\subfigure[using online learning]{
		\includegraphics[width=\linewidth]{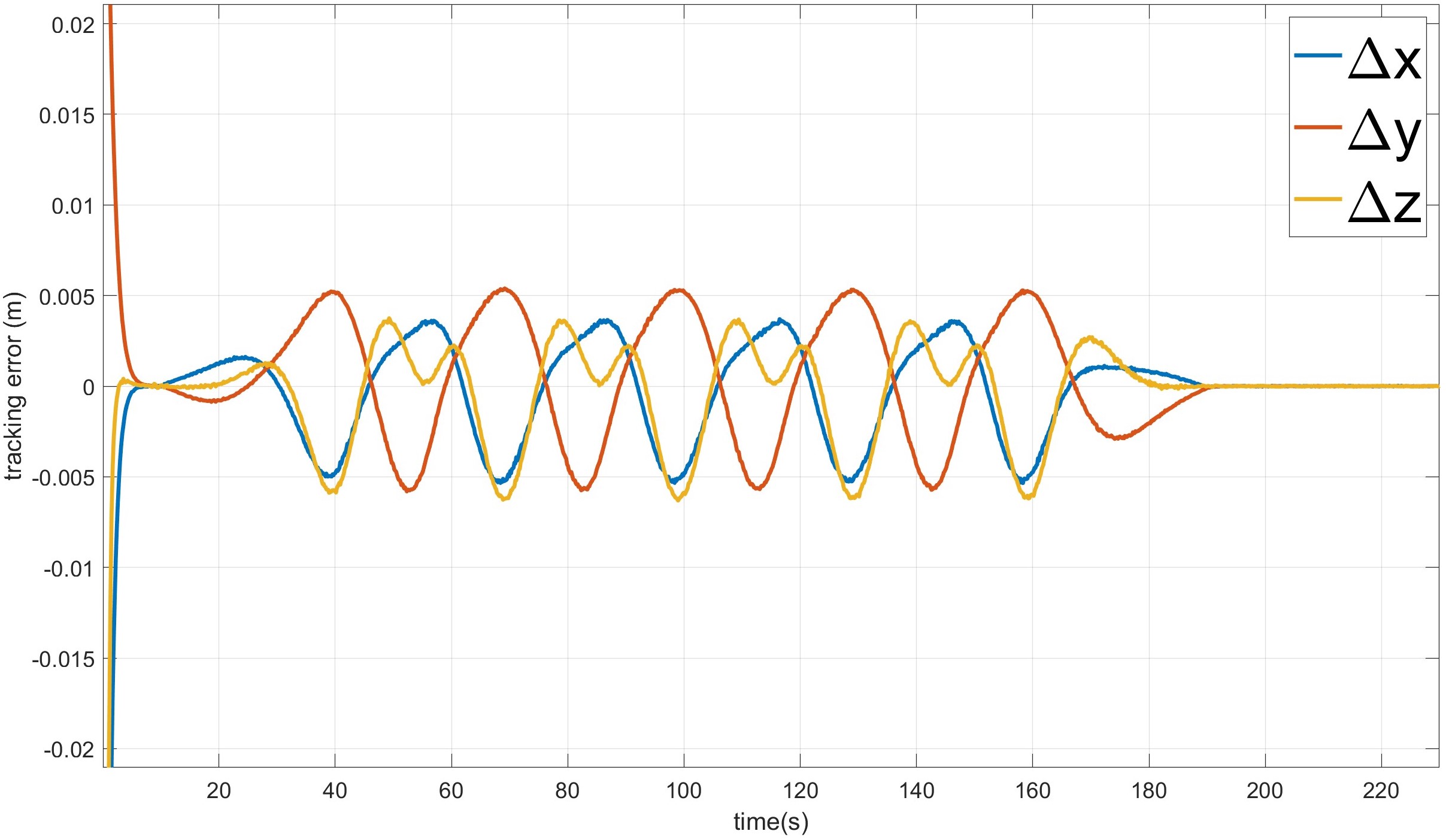}\label{fig:online}}
	\end{minipage}
	\caption{Tracking errors of  kinematic control }
		\begin{minipage}[t]{ \linewidth}
		\centering
			\subfigure[using Leaky ReLU ]{
	    \includegraphics[width=\linewidth]{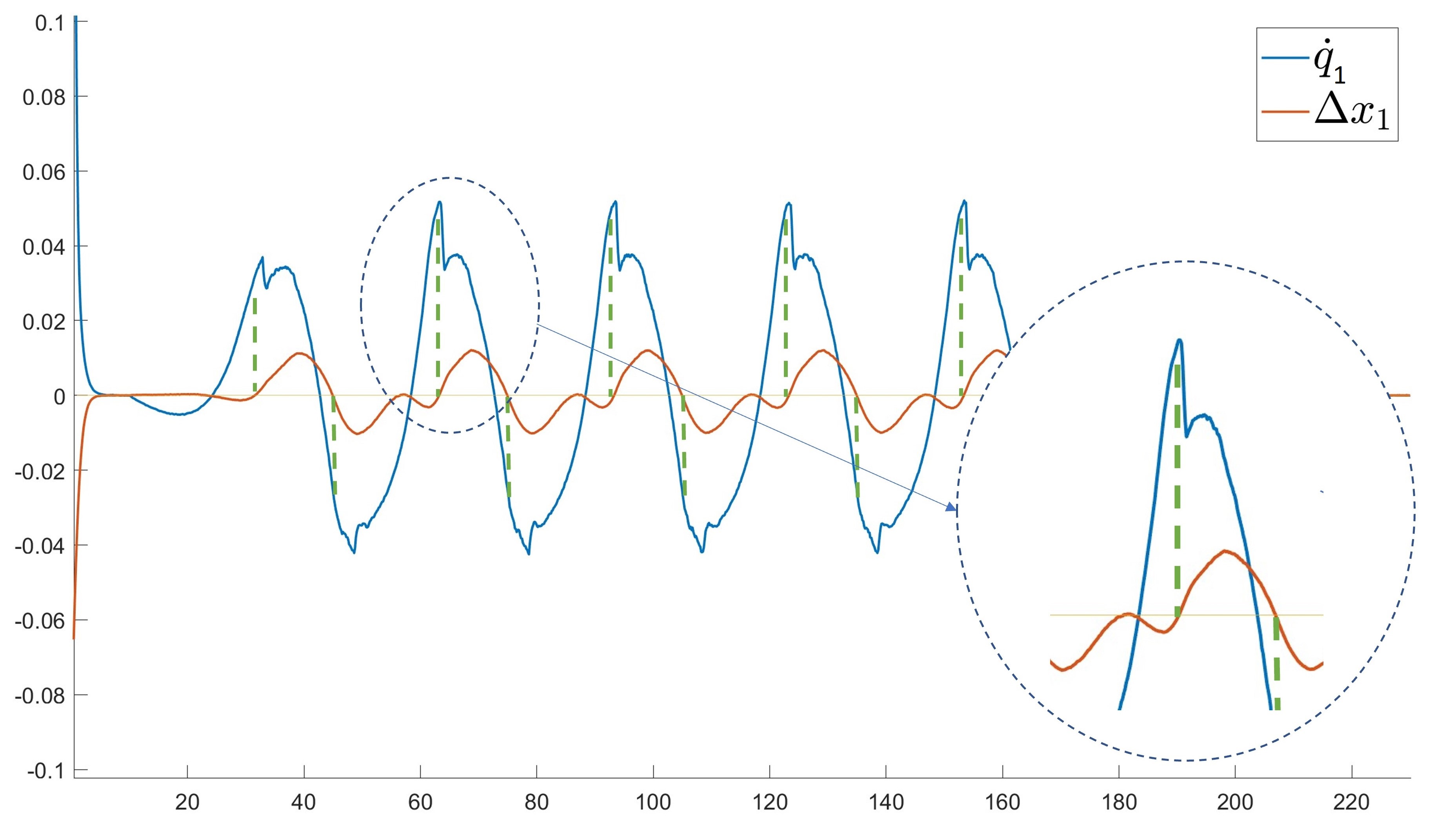}\label{fig:chattering}}
	\end{minipage}
		\begin{minipage}[t]{ \linewidth}
		\centering
			\subfigure[using Sigmoid ]{
	    \includegraphics[width=\linewidth]{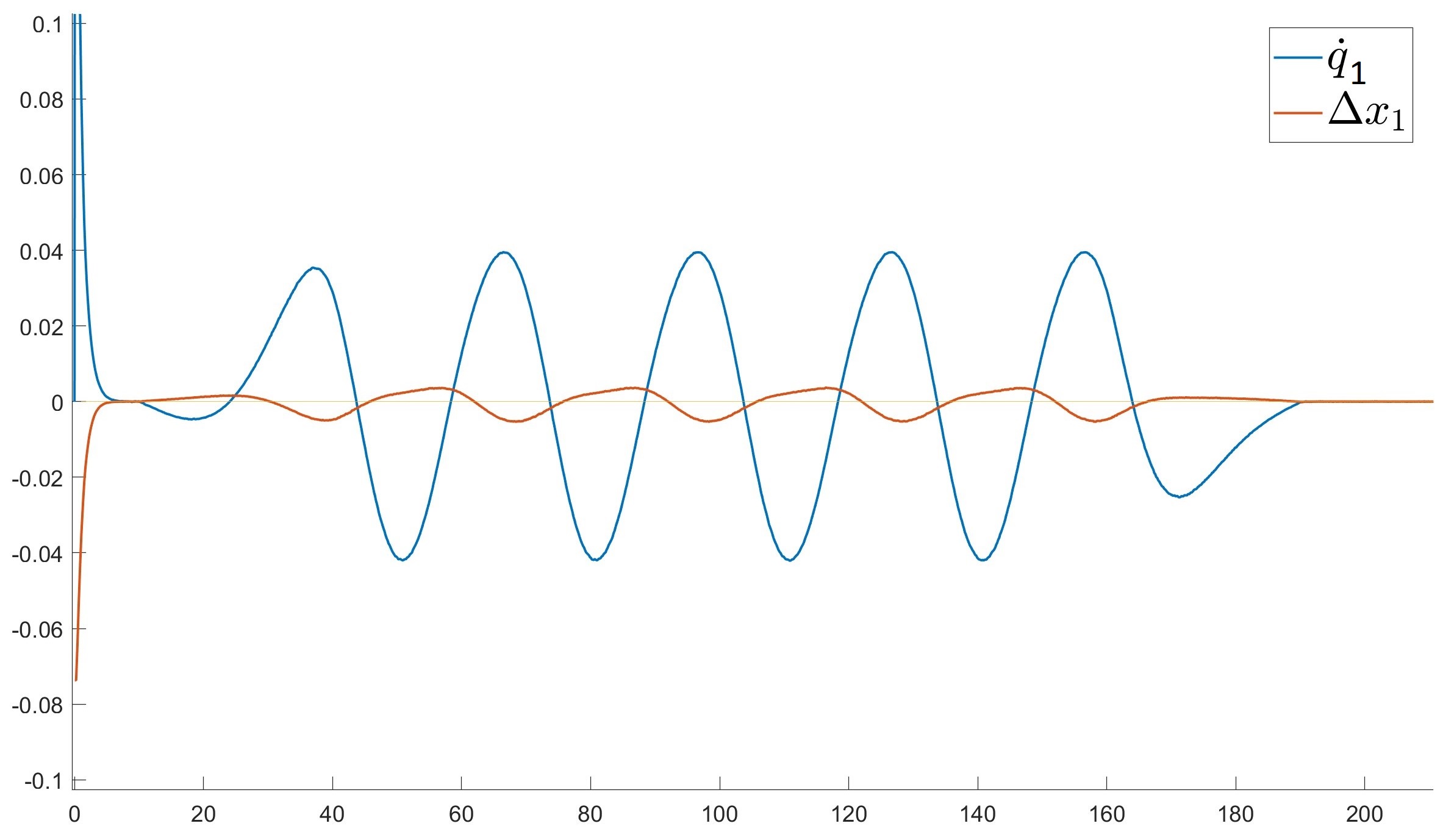}\label{fig:nochattering}}
	\end{minipage}
	\caption{Reference input and tracking error for kinematic control} 
\end{figure}

The online task was to track a circle trajectory   $C1$ with the   parameters shown in Fig \ref{fig:noonline}.
The training data were manually collected around circle $C1$. The pretraining was  conducted by setting $\alpha_j=0$ in \eqref{eq:updL_2layer_b} and $\mathbf{L}_j=\mathbf{L}_n=diag(0.05,...,0.05)$ for 25 epochs and then by setting $\alpha_j^{\vartriangleright}=\alpha_j=\alpha_n=1$ the fine-tuning process was conducted.  The offline  weights were later used as starting weights for the online trajectory tracking task. The tracking errors  are  shown in Fig \ref{fig:noonline}.
%
Then  from the obtained offline weights, the online updating was performed according to \eqref{eq:control_qre}. During the online training, all layers' weights were updated concurrently and only the last sub-system was used for online control. As shown in Fig \ref{fig:online}, after online updating, the tracking errors have been reduced as online learning can further adapt the control system using tracking errors.


The online task was also performed on Leaky ReLU activation function with the same network structure and learning method for comparison purpose. It can be seen from the results in Fig \ref{fig:chattering} that,  Leaky ReLU activation function causes chattering  problem when the error goes through zero due to it is unsmooth when input is zero. However, as shown in Fig \ref{fig:nochattering},  smooth activation function Sigmoid does not have this problem and hence it is differentiable  .
 \section{Conclusion}
 In this paper, an End-to-End deep learning algorithm based on collaborative learning has been proposed. This learning algorithm allows a  deep FNN updating all layer's weight concurrently on $n-1$ subsystems and that the subsystems are combined by a collaborative learning layer. The convergence analysis is provided  with smooth activation functions like Sigmoid to avoid chattering input problem in control tasks. We have shown the approximation ability of the proposed method for  classification tasks compared with SGD, and  the approximation ability for regression task and online kinematics task.

\bibliographystyle{IEEEtran}
\bibliography{reference.bib}

\begin{thebibliography}{10}
\providecommand{\url}[1]{#1}
\csname url@samestyle\endcsname
\providecommand{\newblock}{\relax}
\providecommand{\bibinfo}[2]{#2}
\providecommand{\BIBentrySTDinterwordspacing}{\spaceskip=0pt\relax}
\providecommand{\BIBentryALTinterwordstretchfactor}{4}
\providecommand{\BIBentryALTinterwordspacing}{\spaceskip=\fontdimen2\font plus
\BIBentryALTinterwordstretchfactor\fontdimen3\font minus
  \fontdimen4\font\relax}
\providecommand{\BIBforeignlanguage}[2]{{%
\expandafter\ifx\csname l@#1\endcsname\relax
\typeout{** WARNING: IEEEtran.bst: No hyphenation pattern has been}%
\typeout{** loaded for the language `#1'. Using the pattern for}%
\typeout{** the default language instead.}%
\else
\language=\csname l@#1\endcsname
\fi
#2}}
\providecommand{\BIBdecl}{\relax}
\BIBdecl

\bibitem{lecun2015deep}
Y.~LeCun, Y.~Bengio, and G.~Hinton, ``Deep learning,'' \emph{Nature}, vol. 521,
  no. 7553, pp. 436--444, 2015.

\bibitem{shrestha2019review}
A.~Shrestha and A.~Mahmood, ``Review of deep learning algorithms and
  architectures,'' \emph{IEEE Access}, vol.~7, pp. 53\,040--53\,065, 2019.

\bibitem{zou2020gradient}
D.~Zou, Y.~Cao, D.~Zhou, and Q.~Gu, ``Gradient descent optimizes
  over-parameterized deep relu networks,'' \emph{Machine Learning}, vol. 109,
  no.~3, pp. 467--492, 2020.

\bibitem{du2019gradient}
S.~Du, J.~Lee, H.~Li, L.~Wang, and X.~Zhai, ``Gradient descent finds global
  minima of deep neural networks,'' in \emph{International Conference on
  Machine Learning}.\hskip 1em plus 0.5em minus 0.4em\relax PMLR, 2019, pp.
  1675--1685.

\bibitem{allen2019convergence}
Z.~Allen-Zhu, Y.~Li, and Z.~Song, ``A convergence theory for deep learning via
  over-parameterization,'' in \emph{International Conference on Machine
  Learning}.\hskip 1em plus 0.5em minus 0.4em\relax PMLR, 2019, pp. 242--252.

\bibitem{shin2022effects}
Y.~Shin, ``Effects of depth, width, and initialization: A convergence analysis
  of layer-wise training for deep linear neural networks,'' \emph{Analysis and
  Applications}, vol.~20, no.~01, pp. 73--119, 2022.

\bibitem{nguyen2022analytic}
H.-T. Nguyen, C.~C. Cheah, and K.-A. Toh, ``An analytic layer-wise deep
  learning framework with applications to robotics,'' \emph{Automatica}, vol.
  135, p. 110007, 2022.

\bibitem{nguyen2022layer}
H.-T. Nguyen, S.~Li, and C.~C. Cheah, ``A layer-wise theoretical framework for
  deep learning of convolutional neural networks,'' \emph{IEEE Access},
  vol.~10, pp. 14\,270--14\,287, 2022.

\bibitem{patil2021lyapunov}
O.~S. Patil, D.~M. Le, M.~L. Greene, and W.~E. Dixon, ``Lyapunov-derived
  control and adaptive update laws for inner and outer layer weights of a deep
  neural network,'' \emph{IEEE Control Systems Letters}, vol.~6, pp.
  1855--1860, 2021.

\bibitem{10053845}
S.~Li, H.-T. Nguyen, and C.~C. Cheah, ``A theoretical framework for end-to-end
  learning of deep neural networks with applications to robotics,'' \emph{IEEE
  Access}, vol.~11, pp. 21\,992--22\,006, 2023.

\bibitem{wang2015collaborative}
H.~Wang, N.~Wang, and D.-Y. Yeung, ``Collaborative deep learning for
  recommender systems,'' in \emph{Proceedings of the 21th ACM SIGKDD
  international conference on knowledge discovery and data mining}, 2015, pp.
  1235--1244.

\bibitem{song2018collaborative}
G.~Song and W.~Chai, ``Collaborative learning for deep neural networks,''
  \emph{Advances in neural information processing systems}, vol.~31, 2018.

\bibitem{cheah2015task}
C.~C. Cheah and X.~Li, \emph{Task-space sensory feedback control of robot
  manipulators}.\hskip 1em plus 0.5em minus 0.4em\relax Springer, 2015.

\end{thebibliography}

\appendix
Consider \eqref{eq:deltaV_5} with $k = 0$, we have 
\begin{equation}
\begin{aligned}
&\Delta V(0) =V(1)-V(0)=\\&\resizebox{0.8\linewidth}{!}{$-{\;}2\alpha_j^{\vartriangleright}\sum_{j=1}^{n-1}\boldsymbol{\delta}_j^T(0)\mathbf{L}_j(0)\boldsymbol{e}_j(0) -2\alpha_n\boldsymbol{\delta}_n^T(0)\mathbf{L}_n(0)\boldsymbol{e}_n(0)$}\\
&\resizebox{0.7\linewidth}{!}{${+\;}\boldsymbol{e}_j^T(0) \Big(\sum_{j=1}^{n-1} \alpha_{j}^{{\triangleright}^2} \Vert {\boldsymbol{\hat \sigma}_j}(0) \Vert^2\mathbf{L}_j^T(0)\mathbf{L}_j(0) $} \\&\resizebox{1\linewidth}{!}{$+ \sum_{j=1}^{n-1}\alpha_{j}^{{\vartriangleright}^2}\Vert {\boldsymbol{x}_j}(0) \Vert^2 (\mathbf{L}^T_j(0)\hat{\mathbf{W}}_j^{\triangleright}(0)\mathbf{S}^2_j(0)\hat{\mathbf{W}}_j^{\triangleright T}(0)\mathbf{L}_j(0))\Big) \boldsymbol{e}_j(0)$}
\\& +\boldsymbol{e}_n^T(0)  \alpha_{n}^2 \Vert {\boldsymbol{\hat \phi}_{sub}}(0) \Vert^2\mathbf{L}_n^T(0)\mathbf{L}_n(0)\boldsymbol{e}_n(0)  
\end{aligned}
\end{equation}

When $k=0$, since all elements in $\Delta V\left(k\right)$ are bounded, therefore there exists $d_j$,$d_n$
such that \eqref{djdn} can be satisfied when $k=0$. This means that $\Delta V\left(0\right)<0$ and also means $ \hat{\mathbf{W}}_j\left(1\right)< \hat{\mathbf{W}}_j\left(0\right)$, $ \hat{\mathbf{W}}^\vartriangleright_j\left(1\right)< \hat{\mathbf{W}}_j^\vartriangleright\left(0\right)$. Therefore, by induction, all $ \hat{\mathbf{W}}_j\left(k\right)$ and $ \hat{\mathbf{W}}_j^\vartriangleright\left(k\right)$ bounded. Therefore, for any k, there exists  $d_j$,$d_n$ such that \eqref{djdn} can be satisfied.

\end{document}